\documentclass[letterpaper, 10 pt, conference]{ieeeconf}
\IEEEoverridecommandlockouts 
\usepackage{graphicx, color, amsmath, hyperref}
\usepackage{cite}
\usepackage[official]{eurosym}
\usepackage{siunitx}

\usepackage{fancyhdr}

\usepackage{caption}
\title{\LARGE \bf The Design of Stretch: A Compact, Lightweight Mobile Manipulator for Indoor Human Environments}

\author{Charles C. Kemp, Aaron Edsinger, Henry M. Clever and Blaine Matulevich 
\thanks{Henry M. Clever is with the Georgia Institute of Technology (GT), Atlanta, GA., USA. He was supported in part by NSF GRFP Grant DGE1148903. Aaron Edsinger and Blaine Matulevich are with Hello Robot Inc. (HRI), Martinez, CA, USA. They were supported in part by NIH Award R43AG072982. Charles C. Kemp is with GT and HRI. He owns equity in and works part time for HRI. He and Henry Clever receive royalties from GT for sales of the Stretch RE1 due to a licensing agreement with HRI. \textit{\textbf{Please note that issued and pending patents cover the Stretch RE1.}}}
}
\begin{document}

\renewcommand{\thispagestyle}[1]{}
\pagestyle{fancy}
\renewcommand{\headrulewidth}{0.0pt}
\addtolength{\topmargin}{-1.0cm}
\fancyhead[C]{\textit{Published in peer-reviewed conference (ICRA 2022) with \href{https://doi.org/10.1109/ICRA46639.2022.9811922}{DOI 10.1109{/}ICRA46639.2022.9811922}. Please cite like \cite{icra_version_of_this_paper}.}\vspace{0.6cm}} 

\maketitle

\begin{abstract}
Mobile manipulators for indoor human environments can serve as versatile devices that perform a variety of tasks, yet adoption of this technology has been limited. Reducing size, weight, and cost could facilitate adoption, but risks restricting capabilities. We present a novel design that reduces size, weight, and cost, while supporting a variety of tasks. The core design consists of a two-wheeled differential-drive mobile base, a lift, and a telescoping arm configured to achieve Cartesian motion at the end of the arm. Design extensions include a 1 degree-of-freedom (DOF) wrist to stow a tool, a 2-DOF dexterous wrist to pitch and roll a tool, and a compliant gripper. We justify our design with anthropometry and mathematical models of static stability. We also provide empirical support from teleoperating and autonomously controlling a commercial robot based on our design (the Stretch RE1 from Hello Robot Inc.) to perform tasks in real homes.
\end{abstract}

\section{Introduction}
\label{sec:intro}

Mobile manipulators for indoor human environments have the potential to serve as versatile devices that perform a variety of useful tasks. Examples include assisting people with disabilities~\cite{nguyen2008assistive, wyrobek2008towards, jain2010assistive, hashimoto2013field, chen2013robots, grice2019home}, retrieving and delivering objects~\cite{bohren2011towards, kunze2012searching, sankaran2012failure, kaelbling2013integrated, prakash2013older, pratkanis2013replacing, troniak2013charlie}, cleaning~\cite{cohen2011poop, elliott2017learning, fox2019multi}, organizing~\cite{srinivasa2010herb, abdo2015robot, kazhoyan21easemilestone}, laundry~\cite{maitin2010cloth, miller2012geometric}, exercise~\cite{hu2021robotics}, and entertainment~\cite{pastor2011skill, pr2_theater_2012, mucchiani2020exploring}. To date, mobile manipulators have primarily been used in robotics research labs. Widespread use in homes and offices has yet to be realized, and use in industrial spaces is nascent. 

We posit that reduced size, weight, and cost will improve adoption of this emerging technology. Larger size increases a robot's swept volume, limiting options for collision free navigation and manipulation. Greater mass worsens the consequences of collisions and falls. Larger and heavier robots are more difficult to transport and manually reposition. Higher cost makes applications infeasible and deters production.

\begin{figure}[t!]
\centering
\includegraphics[width=0.65 \columnwidth]{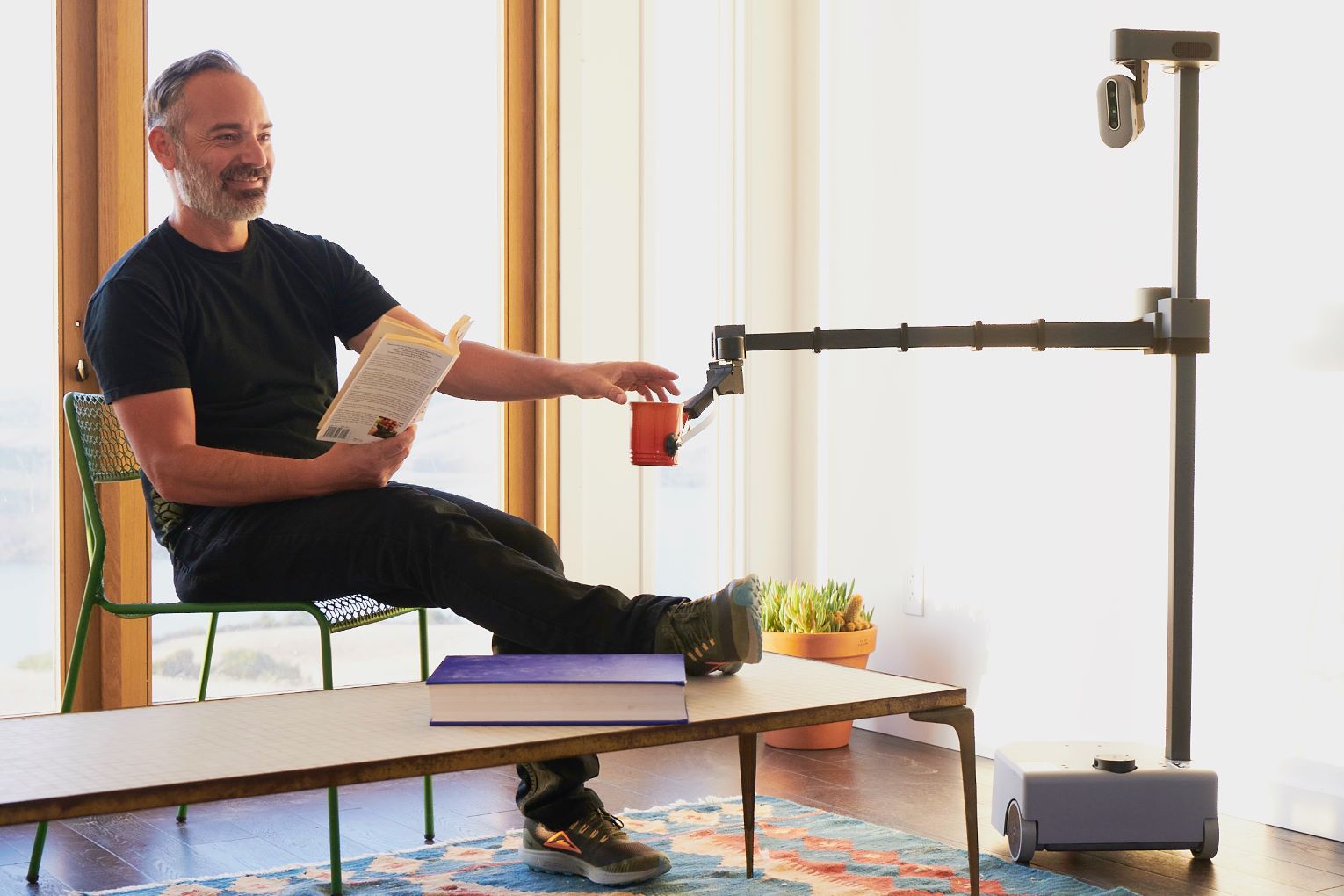}
\caption{\label{fig:intro_image} The Stretch RE1 from Hello Robot Inc. teleoperated to hand an object to Dr. Aaron Edsinger in a real home.}
\vspace{-0.5cm}
\end{figure}

Our primary objective was to create a compact, lightweight, and affordable mobile manipulator capable of performing a variety of useful tasks within indoor human environments. Static stability becomes a dominant concern, since reducing weight and base size reduces the stably achievable workspace and loads \cite{korayem2010maximum}. Reducing the scale of mobile manipulators can make tasks infeasible \cite{kuka_youbot_2011, dusty_2012, ohmm_2013, mersha2018affordable, pinto_home_robot_2018, pyrobot2019}, so we developed a novel design matched to indoor use with better scaling properties \cite{thompson1942growth, mcmahon1983size}. 

To balance competing objectives, we used mechanical models and iterative design. For each iteration, we created a prototype robot and tested it with a variety of real tasks. From October 2016 to July 2017, we created and tested two prototype robots in the Healthcare Robotics Lab at Georgia Tech~\cite{bhattacharjee2018multimodal}. From August 2017 to May 2020, Hello Robot Inc. created a sequence of eight prototype robots with tests in a real home in Atlanta, Georgia, USA.

\begin{table}[h]
  \centering
\resizebox{1.0 \columnwidth}{!}{
  \centering
\begin{tabular}{ | l | l | l | l | }
\hline
\textbf{Robot} & \textbf{Width} & \textbf{Weight} & \textbf{Price} \\
 \hline
 Stretch RE1 by Hello Robot & $34\,\si{\cm}$ & $23\,\si{\kilogram}$ & $\sim$\$20,000\\ 
 HSR by Toyota  \cite{hsr_specs, yamamoto2019development} & $43\,\si{\cm}$ & $37\,\si{\kilogram}$ & N.A.\\
 Fetch by Fetch Robotics \cite{fetch_specs, wise2016fetch} & $51\,\si{\cm}$ & $113\,\si{\kilogram}$ & $\sim$\$100,000 \cite{fetch_price}\\
 TIAGo Steel by PAL Robotics \cite{tiago_specs} & $54\,\si{\cm}$ & $70\,\si{\kilogram}$ & $\sim$\$56,000 \cite{tiago_price}\\
 PR2 by Willow Garage \cite{cousins2014willow, pr2_user_manual_2012, pr2_specs} & $67\,\si{\cm}$ & $220\,\si{\kilogram}$ & \$400,000\\
 \hline
\end{tabular}}
\caption{\label{tab:robots}}
\vspace{-0.3cm}
\end{table}

The resulting product, the Stretch Research Edition 1 (``Stretch RE1'' or ``Stretch''), is significantly smaller, lighter, and less expensive than prior mobile manipulators with comparable capabilities (see Fig. \ref{fig:intro_image} and Table \ref{tab:robots}). Within this paper, we present the design of Stretch, justify it, and provide empirical evidence for its efficacy.

\section{The Design of Stretch}

We created a minimalist design for mobile manipulation in indoor human environments. The Roomba robotic floor cleaner by iRobot served as an inspirational example due to its minimalist design and wide adoption. iRobot began selling the Roomba in 2002 and sold over 1 million robots by late 2004 \cite{irobot2004}. This success was due in part to the Roomba's compact, lightweight, and low-cost design for autonomous floor cleaning in homes. These characteristics were achieved by matching the Roomba's body, sensors, and computation to the task and environment \cite{brooksbuild}.

\subsection{The Structure of Indoor Human Environments}

Indoor human environments have Cartesian structure with horizontal planes and vertical surfaces, including floors, countertops, walls, doors, and cabinet faces. Humans have engineered these environments to facilitate perception, navigation, and manipulation by people. Flat surfaces help objects remain in place. People often rest their bodies on approximately flat surfaces, such as the tops of beds and chair seats. Important locations tend to be accessible via clutter-free paths wide enough to walk through, and important objects tend to be visible from human head heights and reachable by human arms. 

Pets and people, including children, older adults, and people with disabilities, are common occupants of indoor human environments, which creates challenges for safe operation. Falls, pinch points, high velocities, and high accelerations all present hazards. Motions that are difficult to predict can create challenges too, making it harder for pets and people to avoid risky situations. For example, doors and drawers represent common hazards, yet their motions are predictable and people learn to manage the risks. 

\begin{figure}[t!]
\centering
\vspace{1mm}
\includegraphics[width=1.0 \columnwidth]{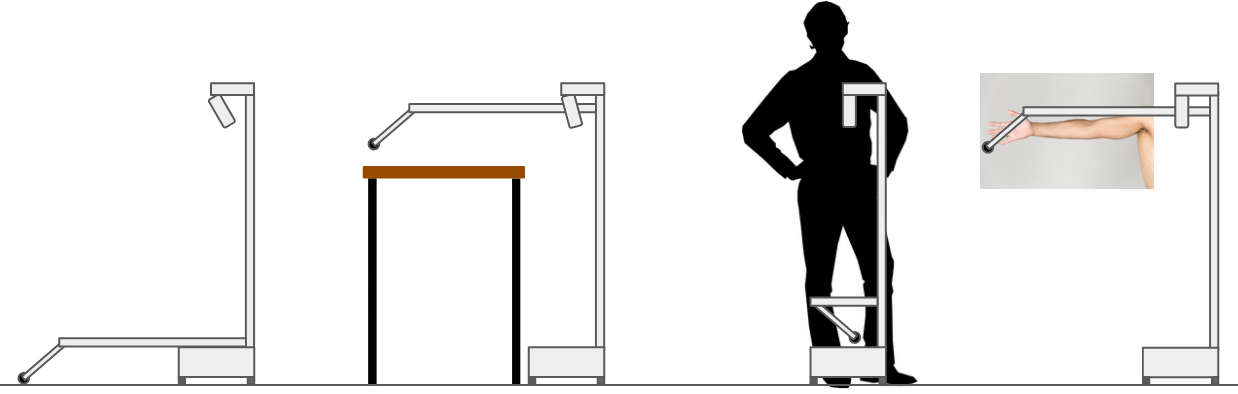}
\vspace{-5.5mm}
\caption{\label{fig:dimensions} Matched to environments \& people (left to right): reach the ground and above countertops; similar to hip width and arm length \cite{human_hips, human_arm}. }
\vspace{-0.5cm}
\end{figure}

Given these considerations, we created a statically-stable wheeled robot to avoid hazards associated with dynamic stability. Stretch weighs $23\,\si{\kilogram}$, which is above safety guidelines for a single person to lift~\cite{hhs1981}, but light enough for two people to lift or a single person to roll around. Our design emphasizes predictable Cartesian motions matched to the Cartesian structure of human environments. Stretch can reach the floor at $0\,\si{\cm}$ up to $110\,\si{\cm}$, enabling manipulation on standard 36-inch high countertops ($\leq 92\,\si{\cm}$) (see Fig. \ref{fig:dimensions}).

\subsection{Matching Human Dimensions}

Using anthropometry, we show that Stretch matches key human dimensions (see Fig. \ref{fig:dimensions}). Rather than being a humanoid root, we think of Stretch as an example of ``robotic cubism'', where the human form has been deconstructed and reassembled in a manner reminiscent of Cubism \cite{wiki:Cubism}. 

When navigating with its arm stowed, Stretch is $34\,\si{\cm}$ wide, which is less than $50^{th}$ percentile hip widths for adult females, $37.1\,\si{\cm}$, and males, $36.1\,\si{\cm}$ \cite{tilley2002measure}. Paths wider than the width of human hips facilitate comfortable walking, and are thus likely to be common in human environments. $34\,\si{\cm}$ is also the diameter of the first Roomba \cite{irobot2002}. 

Frequently manipulated objects tend to be reachable by human arms. The outer edge of Stretch's closed fingertips reaches $71\,\si{\cm}$ out from the edge of the mobile base, which is similar to $50^{th}$ percentile arm lengths for adult females, $67.3\,\si{\cm}$, and males, $72.6\,\si{\cm}$ \cite{tilley2002measure}. 

Frequently observed parts of the environment tend to be visible from human head height. Stretch's color camera is $131\,\si{\cm}$ above the ground, which is between $50^{th}$ percentile sitting eye heights for adult females, $112.3\,\si{\cm}$, and males, $121.9\,\si{\cm}$, and $50^{th}$ percentile standing eye heights for adult females, $151.4\,\si{\cm}$, and males, $164.3\,\si{\cm}$\cite{tilley2002measure}. 

To facilitate assisting people with disabilities, the center of Stretch's closed fingertips reaches $112\,\si{\cm}$ high, which is close to $95^{th}$ percentile right shoulder heights of female, $109.7\, \si{\cm}$, and male, $113.4\, \si{\cm}$, wheelchair users \cite{paquet2004anthropometric}. 

\begin{figure}[t!]
\centering
\vspace{1mm}
\includegraphics[width=1.0 \columnwidth]{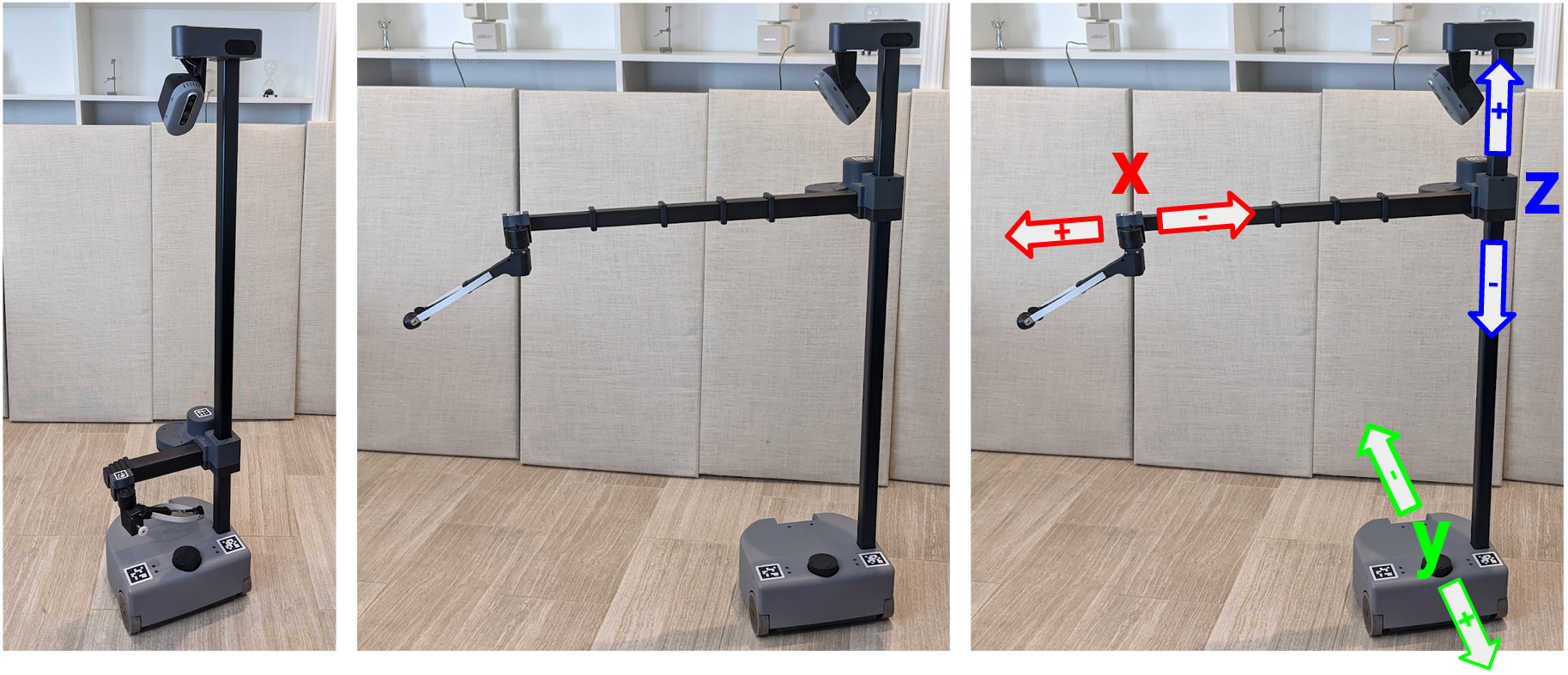}
\vspace{-6.5mm}
\caption{\label{fig:modes} Modes offer insight (left to right): navigation mode; manipulation mode; axes of Cartesian motion for the manipulation mode.}
\vspace{-0.5cm}
\end{figure}

\subsection{The Four Main Joints}

Stretch's core design consists of four actuators that produce Cartesian motion at the end of the arm and mobility across a flat floor (see Fig. \ref{fig:modes}). We prioritized reducing actuator requirements, since actuators influence the weight, complexity, and cost of robots. Each actuator requires power, signals, and control. Conventional electric motors with transmissions add weight, and actuators for proximal joints tend to be heavier. 

One of Stretch's four main actuators extends and retracts a horizontal telescoping arm orthogonal to the mobile base's forward direction of motion. The arm has a small cross section, reducing the mass and swept volume. Its linear motion is readily interpretable, such as when handing an object to a person. Another actuator moves the telescoping arm up and down a vertical mast, which we refer to as the lift. The mast also has a small cross section, which reduces mass and line-of-sight occlusion, and increases reach when the mobile base moves underneath a surface. The remaining two actuators drive the wheels of the differential-drive mobile base to which the mast is mounted. 

Stretch's four main joints represent a minimalist design to achieve Cartesian end-of-arm motion and mobility. Many alternatives have practical limitations. For example, the telescoping arm could have a fixed length and its actuator could be moved to the mobile base to achieve omnidirectional motion using three omniwheels \cite{lynch2017modern}. However, a long arm would increase the robot's width, creating a hazard and limiting movement, while a short arm would lose the benefits of a human-length arm. Adding one or more actuators for omnidirectional base motion would increase the actuator count and could complicate traversal of thresholds and other deviations from flat ground, such as throw rugs, rug fringe, cables, and tile.

To achieve Cartesian motion, the base drives forward and backward ($\pm q_m$) moving the end of the arm left and right ($\pm y_{e}$). The lift ($\pm q_l$) moves the end of the arm up and down ($\pm z_{e}$), and the telescoping arm ($\pm q_a$) moves it out and in ($\pm x_{e}$). Across the full workspace, $[x_{e},y_{e},z_{e}] = [q_a,q_m,q_l]$, the Jacobian is the identity matrix, $I$, and the manipulability ellipsoid is an axis-aligned sphere \cite{craigintroduction}.  

Notably, the mobile base plays a critical role in manipulation, since the lift and telescoping arm only position the end of the arm within a vertical plane. This contrasts with approaches to mobile manipulation that keep a mobile base stationary while manipulating with an arm. Our approach also differs from mounting a conventional serial manipulator to a mobile base. Arms with prismatic joints that extend beyond twice their retracted length are uncommon \cite{mashimo2010analysis, wakita2012user, collins2016design, roman_yim_2018}. 


When operating on flat floors, only the lift actuator works directly against gravity. The telescoping arm uses lightweight materials (e.g., carbon fiber) to achieve a long reach with low weight, reducing the lift actuator's requirements. The arm's lightweight structure also reduces its influence on the robot's center of mass (COM) when it is extended or raised. 

\subsection{Two Modes of Operation}

While they do not represent all achievable motions, it is conceptually helpful to consider two modes of operation: \textit{navigation mode} and \textit{manipulation mode} (see Fig. \ref{fig:modes}). 

\subsubsection{Navigation Mode}

The telescoping arm retracts into the footprint of the mobile base and the robot drives around as a conventional differential-drive mobile robot. The lift can also lower the telescoping arm, lowering the robot's COM and increasing stability. In the navigation mode, the mobile base's forward direction of travel is considered the front of the robot and the direction of focus for sensors, perception, and human-robot interaction. 

\subsubsection{Manipulation Mode}

The front of the robot is the direction the telescoping arm extends, which is orthogonal to the mobile base's forward direction of travel. This mode provides Cartesian motion of the end of the arm. When the mobile base rotates in place, the telescoping arm rotates around a vertical axis, which is similar to rotation with a proximal shoulder joint at the base of a conventional serial manipulator. In addition to pure translations and rotations, the mobile base can perform curvilinear motions. 

\subsubsection{Mode Switching}

Often the robot uses the navigation mode to move to a task-relevant location and then switches to the manipulation mode. A challenge for human operators and autonomous control is to select a good location at which to make this transition. For example, if the goal is to grasp an object, the object should be within reach once the robot switches to the manipulation mode. The robot often rotates in place, lifts its arm, and extends its arm when transitioning from the navigation mode to the manipulation mode. It also rotates its head from looking in the forward direction of travel to looking toward the end-of-arm tool.

\begin{figure}[t!]
\centering
\vspace{1mm}
\includegraphics[width=\columnwidth]{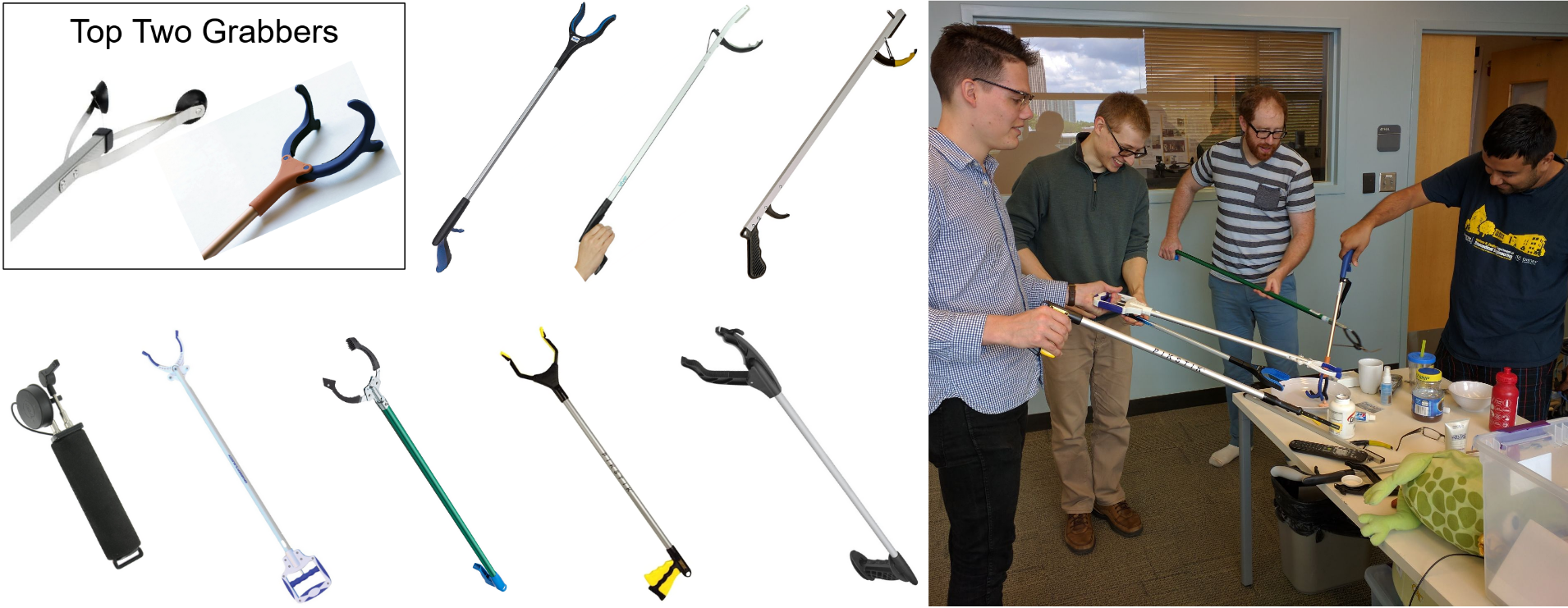}
\vspace{-5.5mm}
\caption{\label{fig:grabbers} \textbf{Left:} Amazon product images for the ten grabbers we evaluated at Georgia Tech in May of 2017. The inset shows the top two. We converted the top left grabber into a robotic gripper. \textbf{Right:} Members of the Healthcare Robotics Lab evaluate grabbers by manipulating objects relevant to assistive robotics, including a pill and a utensil \cite{choi2009list}.}
\vspace{-0.6cm}
\end{figure}

\begin{figure*}[ht!]
\centering
\vspace{1mm}
\includegraphics[width=0.85 \textwidth]{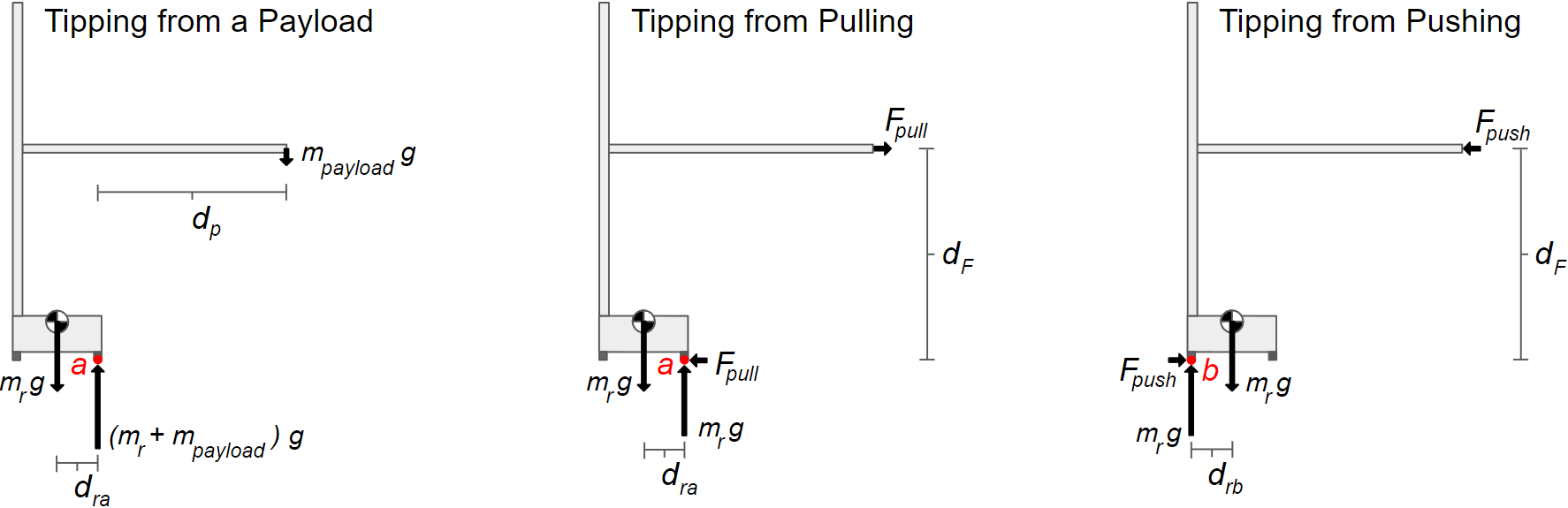}
\vspace{-1.5mm}
\caption{\label{fig:models} Three free body diagrams (FBDs) showing planar models of tipping for task-relevant loads. The robot is in static equilibrium and all but one wheel is losing contact with the ground. Contact with the ground occurs at points $a$ and $b$ (shown in red) on the robot's right and left wheels, respectively.}
\vspace{-0.5cm}
\end{figure*}

\subsection{The Wrist and the Stretch Dex Wrist Accessory}

Stretch comes with a wrist yaw joint with a $330^{\circ}$ range of motion that can stow a tool in the footprint of the mobile base without changing the tool's orientation with respect to gravity. In the navigation mode, the arm retracts and the wrist stows the tool. In the manipulation mode, the wrist swings the tool out, deploying it for use. Stowing tools in this manner makes the effective size of the robot smaller when navigating, avoiding collisions between the tool and the environment. 

In addition to stowing the tool inside the footprint of the base, the wrist yaw joint provides task-relevant dexterity. It is a fifth DOF that adds dexterous redundancy to pure rotations of the mobile base. This is advantageous for manipulation on flat, horizontal surfaces. For example, it can rotate a camera to see behind objects on a shelf and orient the gripper to grasp elongated objects. 

While the yaw joint is sufficient for a variety of tasks, some tasks benefit from additional degrees of freedom. The Stretch Dex Wrist is a recent accessory that adds a pitch joint followed by a roll joint in a serial chain. Together with the yaw joint, this results in a 3-DOF wrist and a 7-DOF robot. 

The Dex Wrist enables tasks such as pouring and operating door knobs (see Fig. \ref{fig:montage}). It also increases the height the center of the closed fingertips can reach to approximately $130\,\si{\cm}$ above the ground, which is close to $95^{th}$ percentile right eye heights of female, $125.5\, \si{\cm}$, and male, $127.0\, \si{\cm}$, wheelchair users \cite{paquet2004anthropometric}. The pitch joint works against gravity, but is a distal joint with a relatively small moment arm. In general, the four main joints provide gross movements over large distances, while the 3-DOF wrist and tools perform fine movements. 

\subsection{The Stretch Compliant Gripper}

In May of 2017 at Georgia Tech, we developed a low-cost compliant gripper (see Fig. \ref{fig:grabbers}). Human-operated grabbers have long been used as affordable assistive devices to help people with disabilities reach and grasp objects in human environments. The grabbers are operated by pulling a cable that can be actuated \cite{song2020grasping}. 

To identify promising candidates, we used Amazon to find and purchase the highest-rated grabbers. Candidates had thousands of ratings and detailed reviews. We purchased ten grabbers ranging from \$13.00 to \$33.95 in price with an average price of \$19.28. 

We then conducted an informal evaluation to select the top candidates. We asked six lab members with expertise in assistive robotics to try the grabbers with objects relevant to assistive robotics, including objects from~\cite{choi2009list} (see Fig. \ref{fig:grabbers}). Based on their feedback, we selected the ``Reacher Grabber by VIVE'' and the ``Japanese Reacher Grabber Pickup Tool''.

We found the rubber fingertips and spring steel flexures of the ``Reacher Grabber by VIVE'' to be highly capable and created a robotic prototype. The commercial version (Stretch Compliant Gripper) that comes with Stretch provides kinematic and torque feedback from the actuator and can exert significantly more grip force than the original human-operated grabber, which is important for tasks such as operating door knobs. It also includes a mount point at a $90^{\circ}$ angle with a hook useful for operating doors and drawers. The standard gripper tilts downwards, which permits grasping objects from above and from the side without use of a pitch joint. The Dex Wrist comes with a different version of this gripper that points straight out, since the Dex Wrist can both pitch and roll the gripper. 

\subsection{Stretch's Sensors}

Stretch can sense applied loads with its actuators using current sensing. The four main joints have low gear ratios that facilitate this form of haptic sensing. The prismatic joints for the lift and arm enable straightforward contact detection, since their non-contact loads remain similar across configurations of the robot. Due to their sensitivity, the robot can autonomously open a a variety of drawers by extending its arm until contacting the drawer's surface, lowering its arm until contacting the handle, and then retracting (see Fig. \ref{fig:montage}).

The robot's RGB-D camera (Intel RealSense D435i) is on a pan-tilt head with a $346^{\circ}$ pan range of motion (ROM) and a $115^{\circ}$ tilt ROM. The head can pan to look in the direction of forward travel while navigating and at the end-of-arm tool while manipulating. It can also look straight down at its mobile base. 

We rotated the camera $90^{\circ}$ from its typical orientation to increase its vertical field of view (i.e., $87^\circ$ for depth and $69^\circ$ for color). This enables it to simultaneously obtain depth imagery from straight ahead and from the floor near the mobile base while navigating. It also gives it a fuller view of the arm's workspace while manipulating. Panning the camera at a fixed downward tilt can capture a useful 3D representation of the surrounding environment, except for the region occluded by the mast. Viewing body-mounted ArUco markers on the mobile base, the wrist, and the proximal part of the arm supports kinematic calibration \cite{munoz2012aruco}. 

The robot has a laser range finder on its mobile base suitable for standard navigation methods. Only the mast occludes the laser range finder when the arm is raised. The robot has a microphone array on the top of the lift for speech recognition and sound source localization. The mobile base also has an inertial measurement unit (IMU) with rate gyros, accelerometers, and a magnetometer that can be used for navigation and tilt detection to avoid tipping and toppling. The wrist has accelerometers that can be used for bump sensing, and the D345i has rate gyros and accelerometers useful for visual perception. 

\vspace{-0.1cm}

\section{Modeling Design Tradeoffs}

Stretch's design results in a fundamental tradeoff between reducing the robot's size and weight and increasing its workspace and applied loads. We used mechanical models to better understand the relationships between size, weight, workspace, and loads. Our analysis benefits from Stretch's design, which is amenable to simple closed-form expressions that represent static stability over the robot's full workspace.

\subsection{Planar Models of Static Stability}

Planar models that represent the robot just before it tips highlight the design tradeoffs. These models assume the robot is in static equilibrium and all but one wheel is losing contact with the ground (see Fig. \ref{fig:models}).

\subsubsection{Maximum Payload}

We first model the payload mass, $m_{payload}$, that results in the robot being in static equilibrium balanced on its right wheel, $point\,a$. The moments about $point\,a$ due to the robot's COM, $m_r g d_{ra}$, and due to the payload, $-m_{payload} g d_p$, must sum to $0$, implying $m_{payload} = m_r \frac{d_{ra}}{d_p}$. 


The maximum payload, $m_{payload}$, is inversely proportional to the reach, $d_p$, proportional to the mass of the robot, $m_r$, and proportional to the width of the mobile base, $d_{ra}$. These relationships highlight key tradeoffs, since we would like higher payloads, but also want the robot to reach farther, weigh less, and be narrower. 

\begin{figure}[t!]
\centering
\includegraphics[width=\columnwidth]{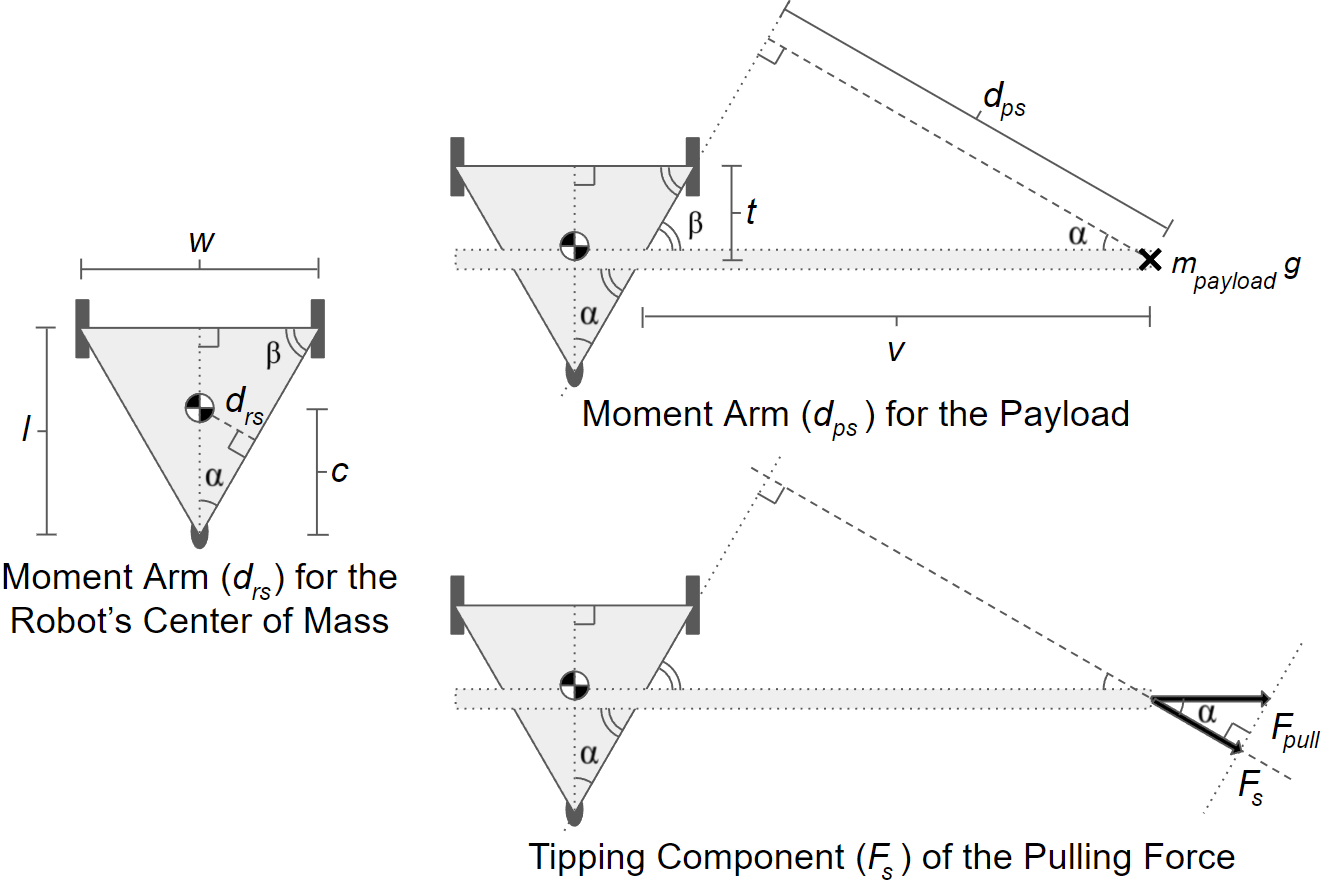}
\vspace{-5mm}
\caption{\label{fig:triangular_base} Stretch's triangular support polygon changes stability under load.}
\vspace{-0.6cm}
\end{figure}

\subsubsection{Maximum Pulling \& Pushing Forces} 

We model the maximum pulling force, $F_{pull}$. The moments about $point\,a$ due to the robot's COM, $m_r g d_{ra}$, and the pulling force, $-F_{pull} d_F$, sum to $0$, implying $F_{pull} = m_r g \frac{d_{ra}}{d_F}$.  

The maximum pulling force is proportional to the mass of the robot, $m_r$, proportional to the width of the mobile base, $d_{ra}$, and inversely proportional to the height of the arm, $d_F$. The robot can pull more strongly when its arm is lower. These relationships highlight additional design tradeoffs, since we would like higher pulling force, but also want the robot to weigh less, be narrower, and reach higher. 

\begin{figure}[t!]
\centering
\includegraphics[width=0.95 \columnwidth]{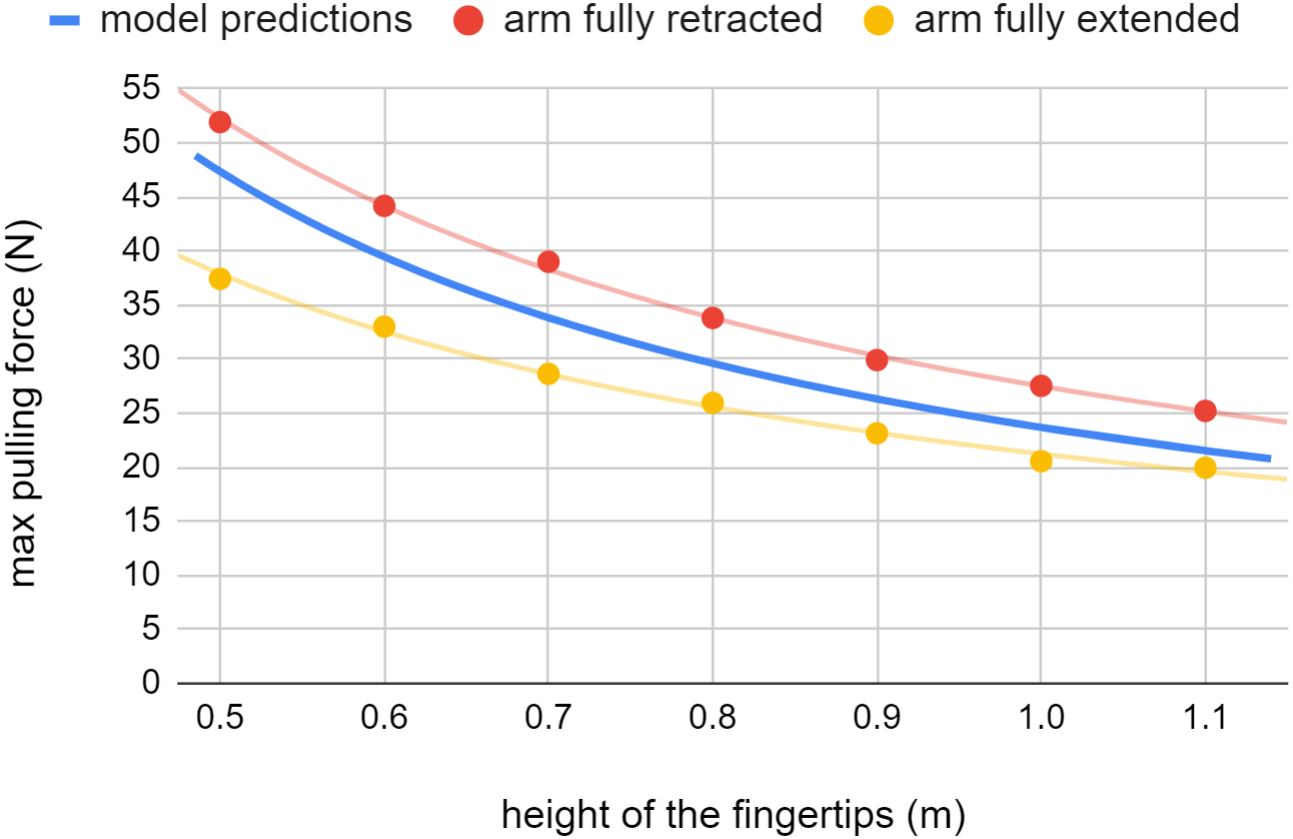}
\vspace{-1mm}
\caption{\label{fig:tipping_force} Pulling force that tips Stretch. Model predictions (\textbf{blue}) and measurements with the arm fully retracted (\textbf{red}) and extended (\textbf{yellow}).}
\vspace{-0.5cm}
\end{figure}

We model the maximum pushing force in a similar way. The moments about $point\,b$ due to the robot's COM, $-m_r g d_{rb}$, and the pushing force, $F_{push} d_F$, sum to $0$, implying $F_{push} = m_r g \frac{d_{rb}}{d_F}$. The main difference between $F_{pull}$ and $F_{push}$ is the moment arm for the robot's COM since $d_{ra} = w - d_{rb}$, where $w$ is the full width of the robot (i.e., the distance between the contact points for the two drive wheels). If the robot's COM is closer to its right wheel, $d_{rb} > d_{ra}$, then $F_{push} > F_{pull}$. If it is closer to its left wheel, $d_{rb} < d_{ra}$, then $F_{push} < F_{pull}$. In practice, the configuration of the telescoping arm changes the lateral position of the robot's COM. For example, fully extending the arm reduces the maximum pulling force and increases the maximum pushing force (see Fig. \ref{fig:tipping_force}). 

\subsection{Static Stability with a Triangular Support Polygon}

Stretch uses a single passive omni wheel to help the drive wheels stay in contact when traversing thresholds and other uneven areas. This is similar to the first Roomba. In practice, Stretch tips over the sides of the support triangle formed by its three wheels.

During development, we updated our planar models to account for this difference (see Fig. \ref{fig:triangular_base}). To simplify analysis, we assume that the robot's COM is equidistant from the two drive wheels. The moment arm of the robot's COM becomes $d_{rs}$, which is the perpendicular distance to the left and right sides of the support triangle. $d_{rs} = c \sin \alpha$, where $c$ is the distance from the robot's COM to the back wheel's point of contact and $\alpha = \arctan \left( \frac{w}{2 l} \right)$, where $w$ is the full width of the robot's support triangle and $l$ is its length. 

The moment arm of the payload changes to $d_{ps}$. $d_{ps} = v \cos \alpha$, where $v = d_p + \frac{t}{\tan \beta}$, $\beta = \frac{\pi}{2} - \alpha$, and $t$ is the distance from the front side of the support triangle to the middle of the telescoping arm. As with our planar models, $d_p$ represents the reach of the telescoping arm beyond the contact point for the right front wheel (see Fig. \ref{fig:models}). Brought together, $d_{ps} = (d_p + \frac{t}{\tan  (\frac{\pi}{2} - \alpha)}) \cos \alpha$.

\begin{figure}[t!]
\centering
\vspace{1mm}
\includegraphics[width=0.95 \columnwidth]{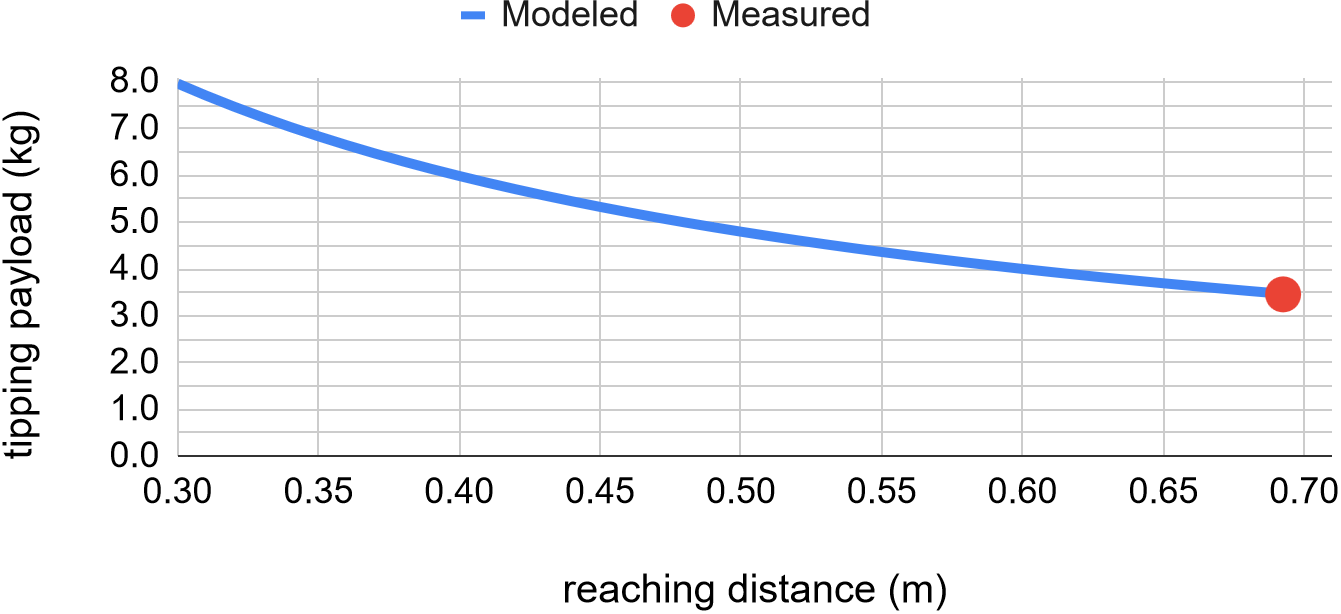}
\vspace{-1mm}
\caption{\label{fig:tipping_payload} Downward load that tips Stretch. \textbf{Blue:} Model predictions. \textbf{Red:} Average measurements for the worst case.}
\end{figure}

\begin{figure*}[ht!]
\centering
\vspace{1mm}
\includegraphics[width=1.0 \textwidth]{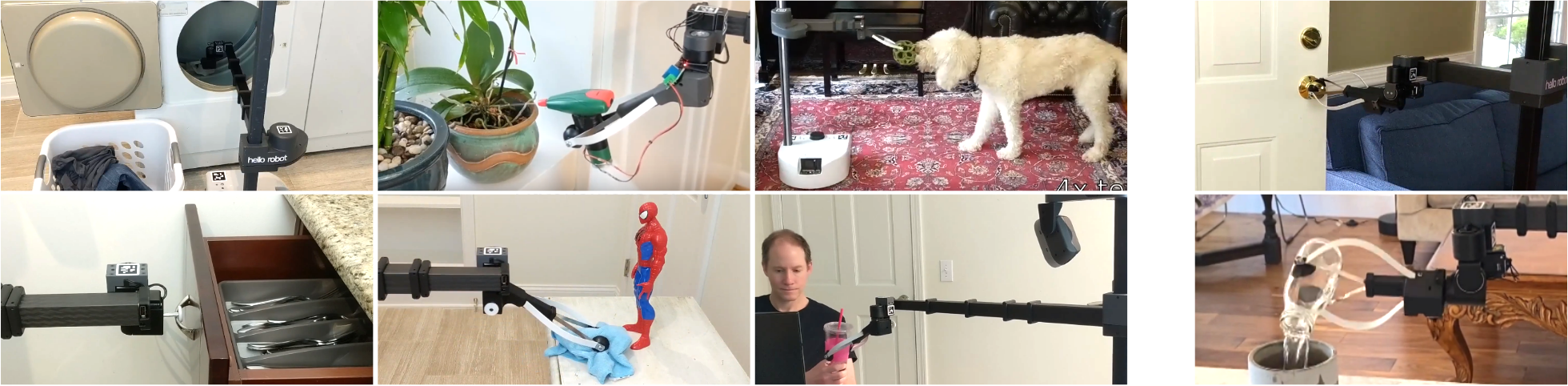}
\caption{\label{fig:montage} Tasks from Table \ref{tab:tasks}. \textbf{Left:} Teleoperated (1, 2, 3) and autonomous (10, 11, 12). \textbf{Right:} Teleoperated with the Stretch Dex Wrist (18, 19).}
\vspace{-0.5cm}
\end{figure*}

For pulling, only the component of the pulling force orthogonal to the side of the support triangle, $F_s$, plays a role, where $F_s = F_{pull} \cos \alpha$. Due to our simplifying assumption for the position of the robot's COM, this model results in the magnitude of the maximum pushing force being the same as the magnitude of the maximum pulling force. As with our planar models, $d_F$ represents the height of the telescoping arm above the floor (see Fig. \ref{fig:models}).
The resulting equations for the maximum payload, $m_{payload}$, and the maximum magnitude of the pulling/pushing force, $F_{pull/push}$, as functions of the reaching distance, $d_p : 0 \leq d_p \leq D$, and height of the telescoping arm, $d_F : 0 \leq d_F \leq H$ follow: 

\vspace{-0.2cm}

\begin{equation}
m_{payload} = m_r \frac{c}{t + \frac{2 l d_p}{w}}    
\end{equation}


\begin{equation}
F_{pull/push} = m_r g \frac{c w}{2 l d_F}    
\end{equation}

$H$, $D$, and $W$ define the height, depth, and width of the robot's Cartesian workspace, $H \times D \times W$, given idealized forward and backward motion of the mobile base. $W$ can be arbitrarily large, since it depends on environmental constraints. We also model the maximum force the robot can apply driving backwards, $F_{backpush} = m_r g \frac{\left( l - c \right)}{d_F}$, based on the height of the arm, $d_F$, and the moment arm for the robot's COM with respect to the front of the support triangle, $l-c$.

\subsection{Other Design Considerations}

A higher COM creates a greater risk of toppling due to acceleration, bumps, thresholds, and ramps. The lightweight aluminum mast, carbon fiber telescoping arm, and small head reduce elevated mass. The batteries and three of the four main actuators are located in the mobile base. When paused, Stretch can be tilted over its drive wheels and rolled around in a manner similar to an upright vacuum cleaner, which can improve ease of use, but also increase the risk of toppling. Ignoring other factors, widening the mobile base by $i\,\si{\cm}$ allows the full extension of the arm to be increased by up to $n\,i \, \si{\cm}$, where $n$ is the number of telescoping elements. 

\subsection{Measurements \& Model Predictions}


We applied our models to a low serial number Stretch RE1 with the default Stretch Compliant Gripper rotated to achieve maximum reach. We used the centers of the wheels for the support triangle and balanced the robot on a metal cylinder to estimate the COM's location giving $w = 0.315\,\si{\meter}$, $l = 0.24\,\si{\meter}$, $m_r = 23 \, \si{\kilogram}$, $g = 9.807 \, \frac{m}{s^2}$, and $c = 0.16 \, \si{\meter}$. Our derivations assumed the load was applied to the middle of the end of the arm. For our evaluation, we applied forces near the fingertips with a Nextech digital force gauge. For this location, $H = 1.125 \, \si{\metre}$, $D = 0.693 \, \si{\meter}$, and $t = 0.005 \, \si{\meter}$. Our models predict that $m_{payload} = \frac{1 \, \si{\kilogram}}{0.0014 + 0.4141 \, \si{\m^{-1}} \, d_p}$,  $F_{pull/push} = \frac{23.68 \, \si{\newton \metre}}{d_F}$ and $F_{backpush} = \frac{18.04 \, \si{\newton \metre}}{d_F}$. 

We teleoperated the robot to a pose, measured the height of contact, $d_F$, and took five pulling force measurements. We used the gauge's hook to pull until the robot began to tip over. For the results in Figure \ref{fig:tipping_force}, each circle shows the average of five measurements. Each group of five measurements has a standard deviation less than $1\,\si{\newton}$. Since our model assumes the COM is equidistant from the left and right sides of the support triangle, we made measurements with the arm fully retracted and fully extended. 

We measured the worst case payload by teleoperating the robot to its maximum reaching distance, $d_{p} = 0.693 \, \si{\meter}$, and pulling down with the gauge's hook until the robot began to tip over (see Fig. \ref{fig:tipping_payload}). The average of five measurements was $3.46\,\si{\kilogram}$ (SD of $0.065\,\si{\kilogram}$), which closely matches our model's $3.47\,\si{\kilogram}$ prediction for $m_{payload}$.

We confirmed that static stability limits the robot's capabilities by fixing the force gauge to the ground near the robot. We disabled the robot's force limits and teleoperated it to pull with $> 70\,\si{\newton}$ and lift with $> 60\,\si{\newton}$ ($> 6.12\,\si{\kilogram}$). To avoid damage, we did not attempt to apply the maximum achievable forces. 

The stably achievable forces and payloads are sufficient for assistive tasks reported in the literature, including opening many drawers and cabinet doors ($< 20\,\si{\newton}$ of pulling force) \cite{jain2010complex, jain2013improving}, face wiping and shaving ($< 10\,\si{\newton}$ of pushing force) \cite{hawkins2012informing}, and lifting objects prioritized for retrieval by people with disabilities ($< 1.2 \, \si{\kilogram}$) \cite{choi2009list}.

\begin{table}[t]
\centering
\vspace{0.2cm}
\resizebox{0.9 \columnwidth}{!}{
  \centering
\begin{tabular}{ | l | l | l | l | }
   \hline
\textbf{\#} & \textbf{Task} & \textbf{Control} & \textbf{Date} \\
 \hline
\textbf{1} &  \textbf{Move laundry from a dryer to a basket} & \textbf{local teleop} & \textbf{Feb 2020}  \\ 
\textbf{2} &  \textbf{Water plants} & \textbf{local teleop} & \textbf{Feb 2020}  \\ 
\textbf{3} &  \textbf{Play with a dog using a ball} & \textbf{local teleop} & \textbf{Feb 2020}  \\
4 &  Vacuum a couch and the floor & local teleop & Feb 2020  \\ 
5 &  Wipe a kitchen countertop & local teleop & Feb 2020  \\  
6 &  Move toys from the floor to a box & local teleop & Feb 2020  \\  
7 &  Pickup pillows and place them on a couch & local teleop & Feb 2020  \\ 
8 &  Hide plastic eggs for kids & local teleop & Feb 2020  \\ 
9 &  Move a toy chicken for kids to shoot & local teleop & Feb 2020  \\ 
\textbf{10} &  \textbf{Pull open drawers using haptic sensing} & \textbf{autonomous} & \textbf{July 2020}  \\ 
\textbf{11} &  \textbf{Wipe a nightstand while avoiding an object} & \textbf{autonomous} & \textbf{July 2020}  \\ 
\textbf{12} &  \textbf{Hand an object to a person} & \textbf{autonomous} & \textbf{July 2020}  \\ 
13 &  Grasp objects from a nightstand & autonomous & June 2020  \\ 
14 &  Write HELLO on a whiteboard & autonomous & July 2020  \\ 
15 &  Map, plan, navigate, and reach a 3D point & autonomous & July 2020  \\  
16 &  Pick and place an object on a shelf & local teleop & July 2020  \\  
17 &  Inspect areas with a camera & local teleop & Dec 2020  \\ 
\textbf{18} & \textbf{Open an exterior door with a door knob} & \textbf{local teleop} & \textbf{Apr 2021}  \\ 
\textbf{19} & \textbf{Pour water and place flowers in a vase} & \textbf{local teleop} & \textbf{June 2021}  \\ 
 \hline
\end{tabular}}
\vspace{-1mm}
\caption{\label{tab:tasks} Tasks Performed by Stretch (\textbf{boldface tasks shown in Fig. \ref{fig:montage})}}
\vspace{-0.5cm}
\end{table}

\section{Testing the Design with a Variety of Tasks}

Table \ref{tab:tasks} lists full tasks we performed with Stretch. Except for the shelf-picking and inspection tasks (16 \& 17), all tasks were in real homes. All of the code and full task videos (see Fig. \ref{fig:montage}) are available on GitHub ({\small \url{https://github.com/hello-robot}}) and YouTube ({\small \url{https://www.youtube.com/c/HelloRobot}}). For local teleoperative control, an expert operator stood by the robot and used a game controller. For autonomous control, the robot performed tasks using only its onboard sensors and computation. 

Other researchers are beginning to disseminate their work with Stretch, providing further evidence for the efficacy of our design. For example, 21 people, including 3 people with disabilities, remotely operated Stretch to perform kitchen tasks \cite{cabrera2021exploration}.

\section{Conclusion}

We presented a novel design for a lightweight, compact, lower cost mobile manipulator capable of performing a variety of tasks in indoor human environments. We supported our design with anthropometry, mechanical models of stability, and tasks in real homes. The growing community using the commercially available Stretch RE1 suggests our design is increasing adoption of mobile manipulators. We are optimistic this community will help create a future in which mobile manipulators improve life for everyone. 

\newpage

\bibliographystyle{IEEEtran}
\bibliography{bibliography}

\end{document}